\pdfoutput=1

\documentclass[11pt]{article}

\usepackage[]{acl}

\usepackage{times}
\usepackage{latexsym}

\usepackage[T1]{fontenc}

\usepackage[utf8]{inputenc}

\usepackage{microtype}
\usepackage{graphicx}
\usepackage{amsmath}  
\usepackage{multirow}
\usepackage{subfigure}

\title{On the Compositional Generalization Gap of In-Context Learning}

\author{Arian Hosseini\\
    Mila, Universit\'e de Montr\'eal \\
     \texttt{arian.hosseini9@gmail.com} \\
    \And Ankit Vani \\
  Mila, Universit\'e de Montr\'eal \\
  \AND
  Dzmitry Bahdanau\\
  ServiceNow Research\\
  \And
  Alessandro Sordoni \\
  Microsoft Research \\
 \And
 Aaron Courville  \\
 Mila, Universit\'e de Montr\'eal \\
 }
\begin{document}
\maketitle
\begin{abstract}
Pretrained large generative language models have shown great performance on many tasks, but exhibit low compositional generalization abilities. Scaling such models has been shown to improve their performance on various NLP tasks even just by conditioning them on a few examples to solve the task without any fine-tuning (also known as in-context learning). In this work, we look at the gap between the in-distribution (ID) and out-of-distribution (OOD) performance of such models in semantic parsing tasks with in-context learning. In the ID settings, the demonstrations are from the same split (\textit{test} or \textit{train}) that the model is being evaluated on, and in the OOD settings, they are from the other split. We look at how the relative generalization gap of in-context learning evolves as models are scaled up. We evaluate four model families, OPT, BLOOM, CodeGen and Codex on three semantic parsing datasets, CFQ, SCAN and GeoQuery with different number of exemplars, and observe a trend of decreasing relative generalization gap as models are scaled up.  
\end{abstract}

\section{Introduction}
Compositional generalization has been a long sought-after goal in deep learning. Typically, when a model is trained on a set of combinations of concepts and tested on novel combinations, it exhibits a lower performance. In contrast, humans excel at combining previously known concepts to generalize to unseen settings. In language, if a human understands the meaning of \emph{green plate}, \emph{black plate} and \emph{green vase}, then they can understand the meaning of \emph{black vase} as well without having seen the combination before.
Big language models have impressive performance on many language understanding tasks \citep{DBLP:conf/naacl/DevlinCLT19, DBLP:journals/jmlr/RaffelSRLNMZLL20, DBLP:journals/corr/abs-2204-02311, lewis-etal-2020-bart}, but they still fail on tasks that require compositional generalization \citep{DBLP:conf/acl/ShawCPT20, DBLP:journals/corr/abs-2007-08970}. 

Prior studies of compositonal generalization use conventional fine-tuning to adapt large language models to the downstream task. The largest recent generative models can be adapted without changing their parameters using \textit{in-context learning}, namely by conditioning them on a prompt with a few exemplars (shots) \citep{DBLP:journals/corr/abs-2204-02311, DBLP:journals/corr/abs-2205-10747, DBLP:conf/nips/BrownMRSKDNSSAA20}. In-context learning benefits particularly well from increased model scale. One can thus wonder whether scaling language models and using them with in-context learning will eventually lead to the disappearance of the compositional generalization gap. 


\begin{figure}[t]
    \centering
    \includegraphics[width=0.5 \textwidth]{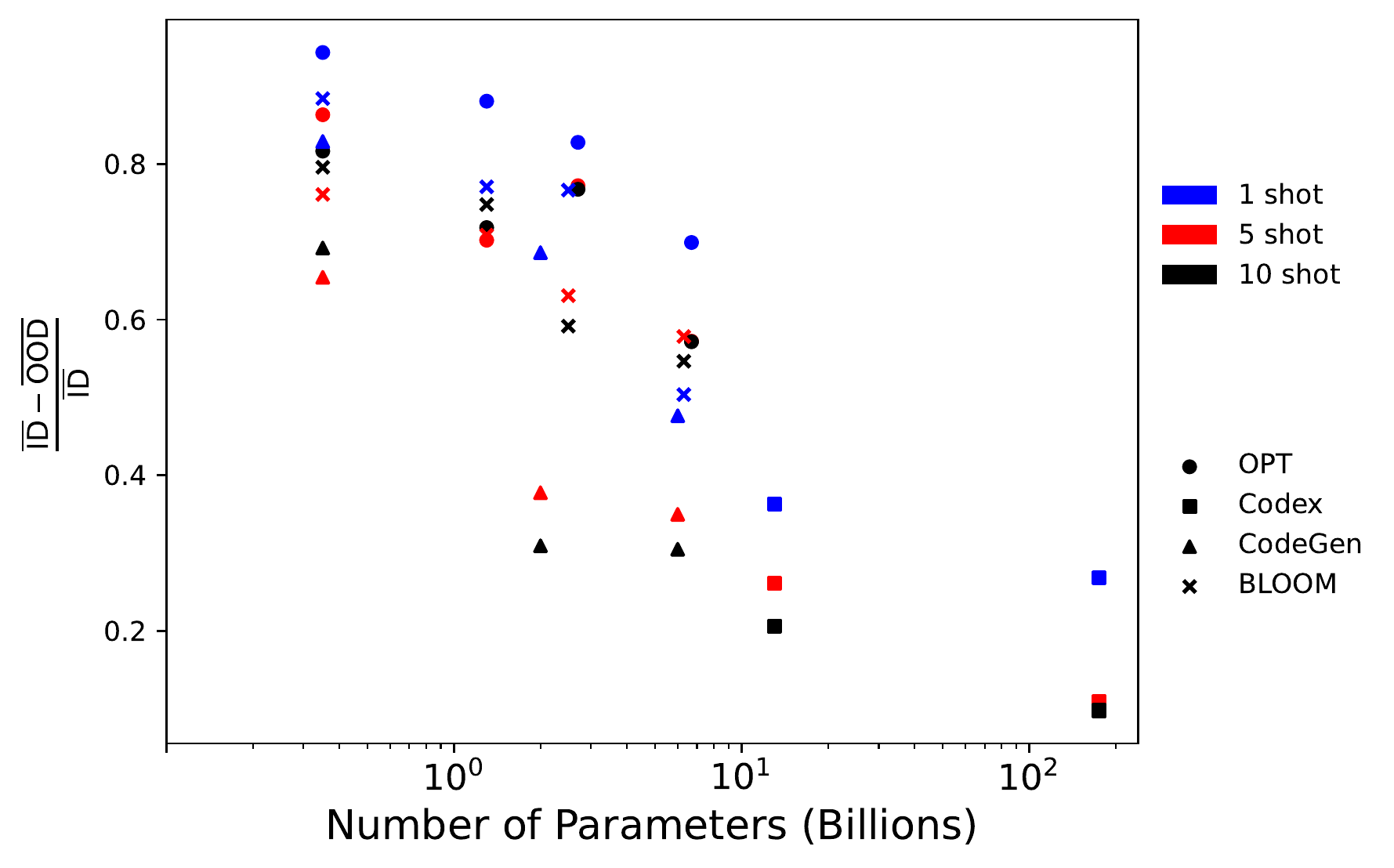}
    \caption{GeoQuery-template relative generalization gap for various models of different sizes across different number of shots. The relative gap is measured by the proportion of in-distribution (ID) performance that is lost when the model receives out-of-distribution (OOD) inputs, ${(\overline{ID} - \overline{OOD})}/{\overline{ID}}$, for each model. Results are averaged over five different seeds.}
    \label{fig:geogengap}
\end{figure}


To answer this question we perform in-context learning experiments on CFQ \citep{DBLP:conf/iclr/KeysersSSBFKMSS20}, SCAN \citep{DBLP:conf/icml/LakeB18}, and GeoQuery \citep{DBLP:conf/aaai/ZelleM96, DBLP:conf/ecml/TangM01} semantic parsing datasets for compositional generalization, and study the generalization gap trend with different number of shots for different models and sizes. Semantic parsing is the task of translating a statement to a logical form with certain syntax and semantics. To solve this task, we provide the model with a prompt constructed of a prefix text and several exemplars from either a split (train or test). Details of constructing the prompt and choosing the exemplars are discussed in section \ref{sec:prompt_design}. We evaluate Codex \citep{DBLP:journals/corr/abs-2107-03374}, BLOOM \citep{bigscience} and CodeGen \citep{Nijkamp2022ACP} which have been pretrained on code as well as natural language. We also evaluate OPT \citep{DBLP:journals/corr/abs-2205-01068} which is only pretrained on natural language data. 

\begin{figure}[t]
    \centering
    \includegraphics[width=0.5 \textwidth]{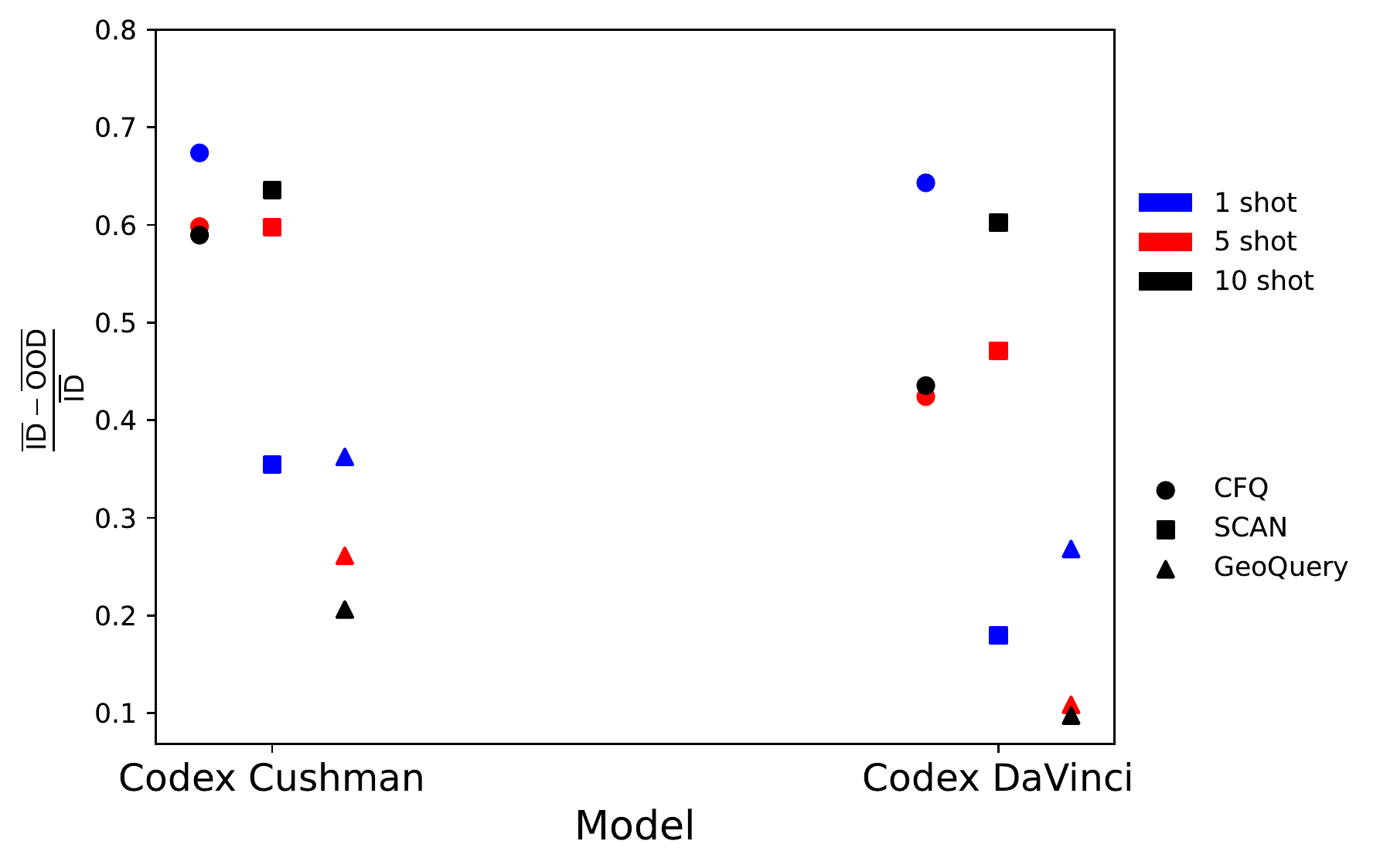}
    \caption{Relative generalization gap on CFQ-MCD1, SCAN-MCD1 and GeoQuery-template for different number of exemplars for Codex DaVinci and Cushman. Results are averaged over five different seeds.}
    \label{fig:cfqgengap}
\end{figure}

We measure how the relative generalization gap of in-context learning evolves as the models are scaled up. We observe a general trend of decreasing relative gap (figure~\ref{fig:geogengap} and figure~\ref{fig:cfqgengap}) as models are scaled up within and across model families with different number of shots.

    


\section{Method}
\label{sec:prompt_design}
For our experiments, we generate prompts that consist of a prefix string introducing the task, followed by a number of exemplars containing inputs and outputs, and finally the test input for which the model will generate an output. 
Inputs and outputs are prefixed with their types, such as ``Command: '' and ``Actions: '' for inputs and outputs respectively in the case of SCAN, and ``Question: '' and ``Query: '' for inputs and outputs respectively in the case of CFQ and GeoQuery. Each input-output pair is separated by an empty line. We refer the reader to Appendix~\ref{app:prompt_design} for the choices of prefix strings and input-output prefixes for each dataset.

We sample our exemplars to maximally cover the primitives in the test input and output. Doing so ensures that our model can use the in-context vocabulary introduced for the specific task rather than using alternative lexicon from its pretrained knowledge.
For natural language inputs, we consider each word as an input primitive. For the formal language outputs, we perform tokenization specific to the language, and consider each token as an output primitive. Note that this tokenization is part of dataset-specific pre-processing and is separate from the tokenization done by the models.

We start selecting exemplars by first greedily collecting successive input-output pairs with the rarest test primitive not already covered by the sampled exemplars. Once the exemplars fully cover the test primitives (in either ID or OOD settings), we sample the remaining exemplars uniformly at random. Table \ref{tab:coverage} shows the coverage percentage of the primitives for different models and datasets. With 10 exemplars, we obtain near-complete primitive coverage for all models and splits.


\begin{figure}[t]
    \centering
    \includegraphics[width=0.50 \textwidth]{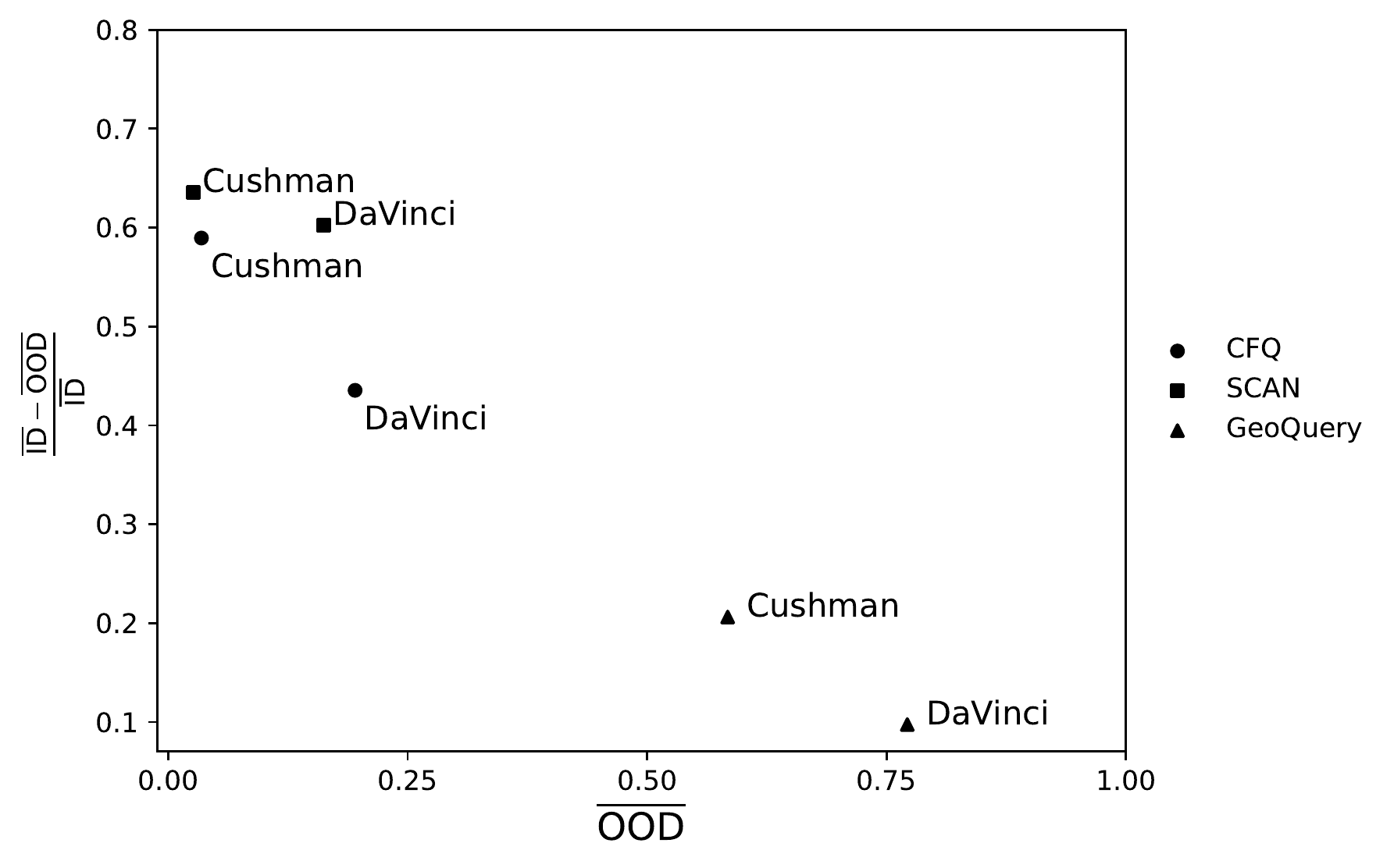}
    \caption{Relative generalization gap with respect to the average OOD generalization performance for Codex DaVinci and Cushman with 10 shots. Ideally, models should be in the lower right corner of this plot. Results are averaged over five different seeds.}
    \label{fig:cfqscanood}
\end{figure}
\section{Experiments}
\label{sec:exps}
We prompt Codex (Cushman and DaVinci), CodGen (350M, 2B and and 6B), OPT (350M, 1.3B, 2.7B and 6.7B) and BLOOM (350M, 1.3B, 2.5B and 6.3B) with queries and exemplars which we sample based on section~\ref{sec:prompt_design} to solve the tasks. We measure and report exact match accuracy for CFQ-MCD1, SCAN-MCD1 and GeoQuery-template subset. Due to execution time constraints of Codex we limited the number of examples to solve to 1045, and compute 95\% confidence interval statistics using 5000 bootstrap samples. Results are averaged over five different seeds which control the sampling of test examples. For CFQ and SCAN, accuracies for models other than Codex are almost zero for all the number of exemplars so we do not include them in our figures and analysis. The models are evaluated on settings defined as $\textbf{split}_A \rightarrow \textbf{split}_B$, which means that the query to be solved is coming from $\textbf{split}_B$, and the exemplars added to the prompt are sampled from $\textbf{split}_A$. We evaluate on four settings: $\textbf{Test} \rightarrow \textbf{Test}$, $\textbf{Train} \rightarrow \textbf{Train}$ which are ID, and $\textbf{Test} \rightarrow \textbf{Train}$, $\textbf{Train} \rightarrow \textbf{Test}$ which are considered OOD. The relative generalization gap is measured as ${(\overline{ID} - \overline{OOD})}/{\overline{ID}}$, where $\overline{ID} =  (Acc(\textrm{Test} \rightarrow \textrm{Test}) + Acc(\textrm{Train} \rightarrow \textrm{Train})) / 2$, and $\overline{OOD} = ( Acc(\textrm{Test} \rightarrow \textrm{Train}) + Acc(\textrm{Train} \rightarrow \textrm{Test})) / 2$.
The relative gap is determined by the proportion of ID performance that is lost when the model receives OOD inputs.

We also plot the relative generalization gap with respect to $\overline{OOD}$ for different tasks and models to get a better understanding of the gap for each model. 
Since higher is better for $\overline{OOD}$, and lower is better for the gap, models closer to the lower right corner of this figure (e.g. figure~\ref{fig:geooodgengap}) are preferred.

\textbf{CFQ} (Compositional Freebase Questions) introduced by \citet{DBLP:conf/iclr/KeysersSSBFKMSS20} is a realistic semantic parsing benchmark to measure compositional generalization. The task is to parse a natural language query, for instance, ``Who directed Elysium'' to a query in SPARQL. We use the MCD-1 (maximum compound divergence) split of CFQ in our experiments. In MCD splits, the authors have maximized the divergence of compound structures and guaranteed low atom divergence between the train and test splits. This makes CFQ an appealing benchmark to measure compositional generalization. We follow the post-processing in \citet{DBLP:journals/corr/abs-2104-07478}, sorting conjuncts alphabetically and deduplicating conjuncts.

\textbf{SCAN} is an instruction following task introduced by \citet{DBLP:conf/icml/LakeB18} where the task is to map natural language instructions (e.g. ``walk thrice'') to action sequences (e.g. ``WALK WALK WALK''). We evaluate Codex DaVinci and Cushman on the MCD-1 split of SCAN. 

\textbf{GeoQuery} is a text-to-SQL dataset \citep{DBLP:conf/aaai/ZelleM96}. We use the \textit{template} split introduced by \citet{DBLP:conf/acl/RadevKZZFRS18} in which train and test splits do not share SQL templates.

\begin{table}
\centering
\small
\begin{tabular}{lcccc}
\hline & \\[-1.5ex]
\multirow{2}{*}{\textbf{Model}} & \multicolumn{2}{c}{$\overline{OOD}$ coverage}  & \multicolumn{2}{c}{$\overline{ID}$ coverage}\\
& 1 shot & 5 shot & 1 shot & 5 shot\\ 
\hline & \\[-1.5ex]
Codex GQ& 75.34\% & 99.91\% & 80.61\% & 99.91\%\\
CodeGen GQ& 75.26\% & 99.91\% & 80.59\% & 99.91\%\\
OPT GQ& 74.69\% & 99.89\% & 80.04\% & 99.92\%\\
BLOOM GQ& 74.78\% & 99.91\% & 80.61\% & 99.88\%\\
\hline & \\[-1.5ex]
Codex CFQ& 54.09\% & 95.81\% & 59.03\% & 98.09\%\\
\hline & \\[-1.5ex]
Codex SCAN& 69.45\% & 100\% & 69.67\% & 100\%\\
\hline
\end{tabular}
\caption{\label{tab:coverage}
Primitive coverage percentage with oracle sampling for GeoQuery-template, CFQ-MCD1 and SCAN-MCD1 splits for Codex, CodeGen, OPT and BLOOM models. The coverage when using 10 shots is 100\% for all models and all splits.
}
\end{table}

\section{Results}

\begin{figure}[t]
    \centering
    \includegraphics[width=0.5\textwidth]{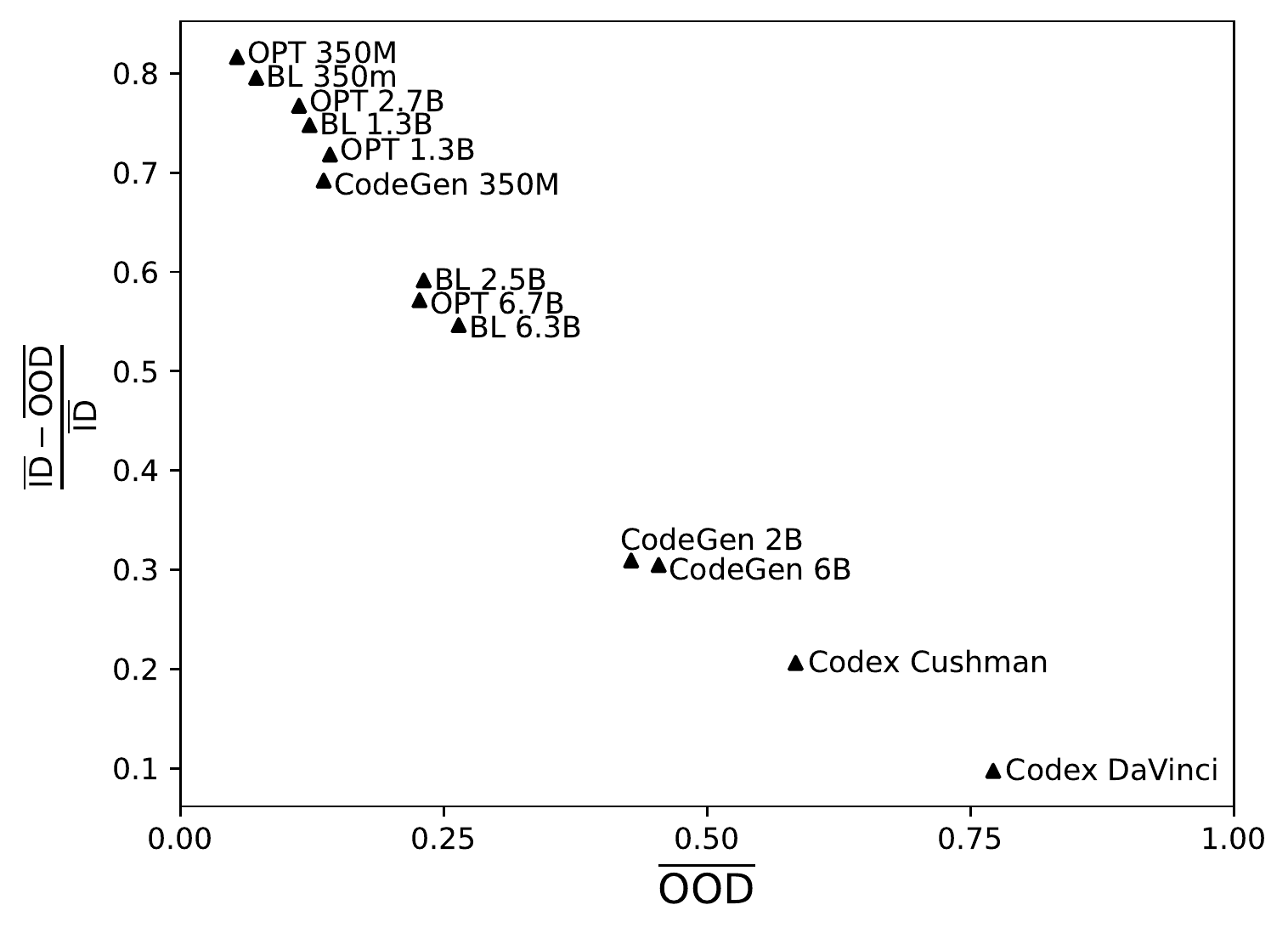}
    \caption{Relative generalization gap with respect to the average OOD generalization performance for GeoQuery-template using 10 exemplars. Ideally, models should be in the lower right corner of this plot. Results are averaged over five different seeds.}
    \label{fig:geooodgengap}
\end{figure}

We study the compositional generalization gap of in-context learning in different large language models of different scale.
Desirable models should perform well OOD and have a low relative generalization gap.
Figure~\ref{fig:geogengap} shows the relative generalization gap for models of different sizes from four model families on the GeoQuery-template dataset for different number of shots. We can observe that the relative generalization gap is smaller for larger models across the four model families. In addition to scale alone, we also find a significant difference in the in-context compositional generalization behavior between different model families.
Particularly, Codex exhibits a higher OOD performance with a low relative generalization gap (see in figure~\ref{fig:geooodgengap}).
Interestingly, Codex is also the only model family out of the ones we considered that achieves ID or OOD performance greater than $1\%$ on CFQ or SCAN. We acknowledge that the two Codex models have the largest amount of parameters amongst the models tested.
Figure~\ref{fig:cfqgengap} shows that as we increase the number of exemplars from 1 to 10 for Codex model family, the relative generalization gap decreases for CFQ and GeoQuery, but increases for SCAN.
In figure~\ref{fig:cfqscanood}, we can see that Codex Cushman generally struggles with both SCAN and CFQ tasks because of the low average OOD generalization score.
It is interesting to note that, for SCAN, Codex DaVinci outperforms Codex Cushman by $\sim$14 points (0.16 vs 0.02) in average OOD generalization performance, albeit their relative generalization gap is similar (as seen in figure~\ref{fig:cfqgengap}). For reference, we report OOD vs. ID performance in appendix \ref{app:ood_v_id}.

We observe a larger set of models performing above near-zero on the GeoQuery dataset, allowing us to compare the generalization gap behavior of other models with increasing scale and number of exemplars. Figure~\ref{fig:geooodgengap} illustrates relative generalization gap with respect to average OOD performance for GeoQuery. In general, we see that models trained on code ~(Codex and CodeGen) are able to achieve higher OOD generalization with lower relative generalization gap on the GeoQuery dataset, with improvements scaling with model size. Since the outputs for GeoQuery dataset contain constructs common in programming languages (appendix~\ref{app:prompt_design}),
these models might have better pretrained knowledge to compositionally generalize to similar tasks with few demonstrations.

\section{Related Work}

Many approaches have tried to improve semantic parsing compositional generalization \citep{DBLP:journals/corr/abs-1904-09708, DBLP:conf/emnlp/LiZWH19, DBLP:conf/iclr/GordonLBB20}.  \citet{DBLP:journals/corr/abs-2104-07478} propose intermediate representations to improve compositional generalization of pretrained seq2seq models. Many have proposed specialized architectures for semantic parsing tasks \citep{DBLP:conf/emnlp/GuptaL18, DBLP:conf/nips/Lake19}. \citet{DBLP:conf/emnlp/ShinLTCRPPKED21} study the adaption of large language models to semantic parsers through few-shot learning. \citet{DBLP:conf/acl/HerzigB20} propose a parser which infers a span tree over the input sequence. The tree specifies how spans are composed together in the input. A line of work studies the use of secondary objectives to improve compositional generalization \citep{DBLP:conf/naacl/YinFNPPSTA21, DBLP:conf/emnlp/JiangB21}.

\citet{DBLP:journals/corr/abs-2007-08970} Study special architectures compared to pretrained language models for semantic parsing. \citet{DBLP:conf/aaai/TsarkovTSMSS21} investigate the compositional generalization abilities of Transformers by scaling the training data size with fixed computational cost.

Large language models are used in different ways to solve downstream tasks. Aside from fine-tuning the model, in-context learning, which is the ability of the model to solve the task by seeing a few exemplars during inference (no weight updates) has gained attention \citep{DBLP:conf/nips/BrownMRSKDNSSAA20, DBLP:conf/icml/WangRHSCBLR22}. Another popular approach, called prompt tuning, is to update a small part of the model's parameters only \citep{DBLP:conf/icml/HoulsbyGJMLGAG19, DBLP:conf/naacl/SchickS21, DBLP:journals/aiopen/HanZDGLHQYZZHHJ21, DBLP:journals/corr/abs-2107-13586, DBLP:journals/corr/abs-2202-04824, DBLP:conf/acl/DingHZCLZS22}. We focus on in-context learning and do not update any parameters. \citet{DBLP:journals/corr/abs-2205-12253} study whether scaling improves compositional generalization in semantic parsing for in-context learning, prompt tuning, and fine-tuning all parameters of the models. We consider their work concurrent to ours with the major difference being that this paper focuses on measuring the relative generalization gap for different model families. As described in detail in section~\ref{sec:exps}, we evaluate on four settings (2 ID and 2 OOD). To the best of our knowledge, \citet{DBLP:journals/corr/abs-2205-12253} only evaluate the $\textbf{Train} \rightarrow \textbf{Test}$ setting.
\section{Conclusion}
We have studied the effect of scaling on the gap between compositional ID and OOD generalization.
We find that the relative generalization gap follows a decreasing trend as models are scaled up for different model families and for different number of support examples.
One factor that limited our study is that in-context learning performance on CFQ and SCAN benchmarks is still very small for almost all publicly available models.
One thing worth investigating in future research is why Codex model family, including the smaller Cushman model, is the only family in this study that achieves above 1\% ID or OOD performance on CFQ or SCAN datasets.
Another interesting future direction is studying the effects of pretraining on code and natural language, rather than natural language alone, on compositional generalization with scaling. Would pretraining on code provide more benefits with increased model scale? Such questions can be answered in the future when the research community has access to more large generative models that are equal in size and amount of training but differ only in data composition. 

\newpage



\bibliographystyle{acl_natbib}
\bibliography{anthology,custom}

\begin{thebibliography}{36}
\expandafter\ifx\csname natexlab\endcsname\relax\def\natexlab#1{#1}\fi

\bibitem[{BigScience(2022)}]{bigscience}
BigScience. 2022.
\newblock BigScience Language Open-science Open-access Multilingual (BLOOM)
  Language Model. International, May 2021-May 2022.

\bibitem[{Brown et~al.(2020)Brown, Mann, Ryder, Subbiah, Kaplan, Dhariwal,
  Neelakantan, Shyam, Sastry, Askell, Agarwal, Herbert{-}Voss, Krueger,
  Henighan, Child, Ramesh, Ziegler, Wu, Winter, Hesse, Chen, Sigler, Litwin,
  Gray, Chess, Clark, Berner, McCandlish, Radford, Sutskever, and
  Amodei}]{DBLP:conf/nips/BrownMRSKDNSSAA20}
Tom~B. Brown, Benjamin Mann, Nick Ryder, Melanie Subbiah, Jared Kaplan,
  Prafulla Dhariwal, Arvind Neelakantan, Pranav Shyam, Girish Sastry, Amanda
  Askell, Sandhini Agarwal, Ariel Herbert{-}Voss, Gretchen Krueger, Tom
  Henighan, Rewon Child, Aditya Ramesh, Daniel~M. Ziegler, Jeffrey Wu, Clemens
  Winter, Christopher Hesse, Mark Chen, Eric Sigler, Mateusz Litwin, Scott
  Gray, Benjamin Chess, Jack Clark, Christopher Berner, Sam McCandlish, Alec
  Radford, Ilya Sutskever, and Dario Amodei. 2020.
\newblock \href
  {https://proceedings.neurips.cc/paper/2020/hash/1457c0d6bfcb4967418bfb8ac142f64a-Abstract.html}
  {Language models are few-shot learners}.
\newblock In \emph{Advances in Neural Information Processing Systems 33: Annual
  Conference on Neural Information Processing Systems 2020, NeurIPS 2020,
  December 6-12, 2020, virtual}.

\bibitem[{Chen et~al.(2021)Chen, Tworek, Jun, Yuan, de~Oliveira~Pinto, Kaplan,
  Edwards, Burda, Joseph, Brockman, Ray, Puri, Krueger, Petrov, Khlaaf, Sastry,
  Mishkin, Chan, Gray, Ryder, Pavlov, Power, Kaiser, Bavarian, Winter, Tillet,
  Such, Cummings, Plappert, Chantzis, Barnes, Herbert{-}Voss, Guss, Nichol,
  Paino, Tezak, Tang, Babuschkin, Balaji, Jain, Saunders, Hesse, Carr, Leike,
  Achiam, Misra, Morikawa, Radford, Knight, Brundage, Murati, Mayer, Welinder,
  McGrew, Amodei, McCandlish, Sutskever, and
  Zaremba}]{DBLP:journals/corr/abs-2107-03374}
Mark Chen, Jerry Tworek, Heewoo Jun, Qiming Yuan, Henrique~Ponde
  de~Oliveira~Pinto, Jared Kaplan, Harrison Edwards, Yuri Burda, Nicholas
  Joseph, Greg Brockman, Alex Ray, Raul Puri, Gretchen Krueger, Michael Petrov,
  Heidy Khlaaf, Girish Sastry, Pamela Mishkin, Brooke Chan, Scott Gray, Nick
  Ryder, Mikhail Pavlov, Alethea Power, Lukasz Kaiser, Mohammad Bavarian,
  Clemens Winter, Philippe Tillet, Felipe~Petroski Such, Dave Cummings,
  Matthias Plappert, Fotios Chantzis, Elizabeth Barnes, Ariel Herbert{-}Voss,
  William~Hebgen Guss, Alex Nichol, Alex Paino, Nikolas Tezak, Jie Tang, Igor
  Babuschkin, Suchir Balaji, Shantanu Jain, William Saunders, Christopher
  Hesse, Andrew~N. Carr, Jan Leike, Joshua Achiam, Vedant Misra, Evan Morikawa,
  Alec Radford, Matthew Knight, Miles Brundage, Mira Murati, Katie Mayer, Peter
  Welinder, Bob McGrew, Dario Amodei, Sam McCandlish, Ilya Sutskever, and
  Wojciech Zaremba. 2021.
\newblock \href {http://arxiv.org/abs/2107.03374} {Evaluating large language
  models trained on code}.
\newblock \emph{CoRR}, abs/2107.03374.

\bibitem[{Chen et~al.(2022)Chen, Liu, Dong, Wang, Zhu, Zeng, and
  Zhang}]{DBLP:journals/corr/abs-2202-04824}
Yulong Chen, Yang Liu, Li~Dong, Shuohang Wang, Chenguang Zhu, Michael Zeng, and
  Yue Zhang. 2022.
\newblock \href {http://arxiv.org/abs/2202.04824} {Adaprompt: Adaptive model
  training for prompt-based {NLP}}.
\newblock \emph{CoRR}, abs/2202.04824.

\bibitem[{Chowdhery et~al.(2022)Chowdhery, Narang, Devlin, Bosma, Mishra,
  Roberts, Barham, Chung, Sutton, Gehrmann, Schuh, Shi, Tsvyashchenko, Maynez,
  Rao, Barnes, Tay, Shazeer, Prabhakaran, Reif, Du, Hutchinson, Pope, Bradbury,
  Austin, Isard, Gur{-}Ari, Yin, Duke, Levskaya, Ghemawat, Dev, Michalewski,
  Garcia, Misra, Robinson, Fedus, Zhou, Ippolito, Luan, Lim, Zoph, Spiridonov,
  Sepassi, Dohan, Agrawal, Omernick, Dai, Pillai, Pellat, Lewkowycz, Moreira,
  Child, Polozov, Lee, Zhou, Wang, Saeta, Diaz, Firat, Catasta, Wei,
  Meier{-}Hellstern, Eck, Dean, Petrov, and
  Fiedel}]{DBLP:journals/corr/abs-2204-02311}
Aakanksha Chowdhery, Sharan Narang, Jacob Devlin, Maarten Bosma, Gaurav Mishra,
  Adam Roberts, Paul Barham, Hyung~Won Chung, Charles Sutton, Sebastian
  Gehrmann, Parker Schuh, Kensen Shi, Sasha Tsvyashchenko, Joshua Maynez,
  Abhishek Rao, Parker Barnes, Yi~Tay, Noam Shazeer, Vinodkumar Prabhakaran,
  Emily Reif, Nan Du, Ben Hutchinson, Reiner Pope, James Bradbury, Jacob
  Austin, Michael Isard, Guy Gur{-}Ari, Pengcheng Yin, Toju Duke, Anselm
  Levskaya, Sanjay Ghemawat, Sunipa Dev, Henryk Michalewski, Xavier Garcia,
  Vedant Misra, Kevin Robinson, Liam Fedus, Denny Zhou, Daphne Ippolito, David
  Luan, Hyeontaek Lim, Barret Zoph, Alexander Spiridonov, Ryan Sepassi, David
  Dohan, Shivani Agrawal, Mark Omernick, Andrew~M. Dai,
  Thanumalayan~Sankaranarayana Pillai, Marie Pellat, Aitor Lewkowycz, Erica
  Moreira, Rewon Child, Oleksandr Polozov, Katherine Lee, Zongwei Zhou, Xuezhi
  Wang, Brennan Saeta, Mark Diaz, Orhan Firat, Michele Catasta, Jason Wei,
  Kathy Meier{-}Hellstern, Douglas Eck, Jeff Dean, Slav Petrov, and Noah
  Fiedel. 2022.
\newblock \href {https://doi.org/10.48550/arXiv.2204.02311} {Palm: Scaling
  language modeling with pathways}.
\newblock \emph{CoRR}, abs/2204.02311.

\bibitem[{Devlin et~al.(2019)Devlin, Chang, Lee, and
  Toutanova}]{DBLP:conf/naacl/DevlinCLT19}
Jacob Devlin, Ming{-}Wei Chang, Kenton Lee, and Kristina Toutanova. 2019.
\newblock \href {https://doi.org/10.18653/v1/n19-1423} {{BERT:} pre-training of
  deep bidirectional transformers for language understanding}.
\newblock In \emph{Proceedings of the 2019 Conference of the North American
  Chapter of the Association for Computational Linguistics: Human Language
  Technologies, {NAACL-HLT} 2019, Minneapolis, MN, USA, June 2-7, 2019, Volume
  1 (Long and Short Papers)}, pages 4171--4186. Association for Computational
  Linguistics.

\bibitem[{Ding et~al.(2022)Ding, Hu, Zhao, Chen, Liu, Zheng, and
  Sun}]{DBLP:conf/acl/DingHZCLZS22}
Ning Ding, Shengding Hu, Weilin Zhao, Yulin Chen, Zhiyuan Liu, Haitao Zheng,
  and Maosong Sun. 2022.
\newblock \href {https://aclanthology.org/2022.acl-demo.10} {Openprompt: An
  open-source framework for prompt-learning}.
\newblock In \emph{Proceedings of the 60th Annual Meeting of the Association
  for Computational Linguistics, {ACL} 2022 - System Demonstrations, Dublin,
  Ireland, May 22-27, 2022}, pages 105--113. Association for Computational
  Linguistics.

\bibitem[{Finegan{-}Dollak et~al.(2018)Finegan{-}Dollak, Kummerfeld, Zhang,
  Ramanathan, Sadasivam, Zhang, and Radev}]{DBLP:conf/acl/RadevKZZFRS18}
Catherine Finegan{-}Dollak, Jonathan~K. Kummerfeld, Li~Zhang, Karthik
  Ramanathan, Sesh Sadasivam, Rui Zhang, and Dragomir~R. Radev. 2018.
\newblock \href {https://doi.org/10.18653/v1/P18-1033} {Improving text-to-sql
  evaluation methodology}.
\newblock In \emph{Proceedings of the 56th Annual Meeting of the Association
  for Computational Linguistics, {ACL} 2018, Melbourne, Australia, July 15-20,
  2018, Volume 1: Long Papers}, pages 351--360. Association for Computational
  Linguistics.

\bibitem[{Furrer et~al.(2020)Furrer, van Zee, Scales, and
  Sch{\"{a}}rli}]{DBLP:journals/corr/abs-2007-08970}
Daniel Furrer, Marc van Zee, Nathan Scales, and Nathanael Sch{\"{a}}rli. 2020.
\newblock \href {http://arxiv.org/abs/2007.08970} {Compositional generalization
  in semantic parsing: Pre-training vs. specialized architectures}.
\newblock \emph{CoRR}, abs/2007.08970.

\bibitem[{Gordon et~al.(2020)Gordon, Lopez{-}Paz, Baroni, and
  Bouchacourt}]{DBLP:conf/iclr/GordonLBB20}
Jonathan Gordon, David Lopez{-}Paz, Marco Baroni, and Diane Bouchacourt. 2020.
\newblock \href {https://openreview.net/forum?id=SylVNerFvr} {Permutation
  equivariant models for compositional generalization in language}.
\newblock In \emph{8th International Conference on Learning Representations,
  {ICLR} 2020, Addis Ababa, Ethiopia, April 26-30, 2020}. OpenReview.net.

\bibitem[{Gupta and Lewis(2018)}]{DBLP:conf/emnlp/GuptaL18}
Nitish Gupta and Mike Lewis. 2018.
\newblock \href {https://doi.org/10.18653/v1/d18-1239} {Neural compositional
  denotational semantics for question answering}.
\newblock In \emph{Proceedings of the 2018 Conference on Empirical Methods in
  Natural Language Processing, Brussels, Belgium, October 31 - November 4,
  2018}, pages 2152--2161. Association for Computational Linguistics.

\bibitem[{Han et~al.(2021)Han, Zhang, Ding, Gu, Liu, Huo, Qiu, Yao, Zhang,
  Zhang, Han, Huang, Jin, Lan, Liu, Liu, Lu, Qiu, Song, Tang, Wen, Yuan, Zhao,
  and Zhu}]{DBLP:journals/aiopen/HanZDGLHQYZZHHJ21}
Xu~Han, Zhengyan Zhang, Ning Ding, Yuxian Gu, Xiao Liu, Yuqi Huo, Jiezhong Qiu,
  Yuan Yao, Ao~Zhang, Liang Zhang, Wentao Han, Minlie Huang, Qin Jin, Yanyan
  Lan, Yang Liu, Zhiyuan Liu, Zhiwu Lu, Xipeng Qiu, Ruihua Song, Jie Tang,
  Ji{-}Rong Wen, Jinhui Yuan, Wayne~Xin Zhao, and Jun Zhu. 2021.
\newblock \href {https://doi.org/10.1016/j.aiopen.2021.08.002} {Pre-trained
  models: Past, present and future}.
\newblock \emph{{AI} Open}, 2:225--250.

\bibitem[{Herzig and Berant(2021)}]{DBLP:conf/acl/HerzigB20}
Jonathan Herzig and Jonathan Berant. 2021.
\newblock \href {https://doi.org/10.18653/v1/2021.acl-long.74} {Span-based
  semantic parsing for compositional generalization}.
\newblock In \emph{Proceedings of the 59th Annual Meeting of the Association
  for Computational Linguistics and the 11th International Joint Conference on
  Natural Language Processing, {ACL/IJCNLP} 2021, (Volume 1: Long Papers),
  Virtual Event, August 1-6, 2021}, pages 908--921. Association for
  Computational Linguistics.

\bibitem[{Herzig et~al.(2021)Herzig, Shaw, Chang, Guu, Pasupat, and
  Zhang}]{DBLP:journals/corr/abs-2104-07478}
Jonathan Herzig, Peter Shaw, Ming{-}Wei Chang, Kelvin Guu, Panupong Pasupat,
  and Yuan Zhang. 2021.
\newblock \href {http://arxiv.org/abs/2104.07478} {Unlocking compositional
  generalization in pre-trained models using intermediate representations}.
\newblock \emph{CoRR}, abs/2104.07478.

\bibitem[{Houlsby et~al.(2019)Houlsby, Giurgiu, Jastrzebski, Morrone,
  de~Laroussilhe, Gesmundo, Attariyan, and
  Gelly}]{DBLP:conf/icml/HoulsbyGJMLGAG19}
Neil Houlsby, Andrei Giurgiu, Stanislaw Jastrzebski, Bruna Morrone, Quentin
  de~Laroussilhe, Andrea Gesmundo, Mona Attariyan, and Sylvain Gelly. 2019.
\newblock \href {http://proceedings.mlr.press/v97/houlsby19a.html}
  {Parameter-efficient transfer learning for {NLP}}.
\newblock In \emph{Proceedings of the 36th International Conference on Machine
  Learning, {ICML} 2019, 9-15 June 2019, Long Beach, California, {USA}},
  volume~97 of \emph{Proceedings of Machine Learning Research}, pages
  2790--2799. {PMLR}.

\bibitem[{Jiang and Bansal(2021)}]{DBLP:conf/emnlp/JiangB21}
Yichen Jiang and Mohit Bansal. 2021.
\newblock \href {https://doi.org/10.18653/v1/2021.emnlp-main.505} {Inducing
  transformer's compositional generalization ability via auxiliary sequence
  prediction tasks}.
\newblock In \emph{Proceedings of the 2021 Conference on Empirical Methods in
  Natural Language Processing, {EMNLP} 2021, Virtual Event / Punta Cana,
  Dominican Republic, 7-11 November, 2021}, pages 6253--6265. Association for
  Computational Linguistics.

\bibitem[{Keysers et~al.(2020)Keysers, Sch{\"{a}}rli, Scales, Buisman, Furrer,
  Kashubin, Momchev, Sinopalnikov, Stafiniak, Tihon, Tsarkov, Wang, van Zee,
  and Bousquet}]{DBLP:conf/iclr/KeysersSSBFKMSS20}
Daniel Keysers, Nathanael Sch{\"{a}}rli, Nathan Scales, Hylke Buisman, Daniel
  Furrer, Sergii Kashubin, Nikola Momchev, Danila Sinopalnikov, Lukasz
  Stafiniak, Tibor Tihon, Dmitry Tsarkov, Xiao Wang, Marc van Zee, and Olivier
  Bousquet. 2020.
\newblock \href {https://openreview.net/forum?id=SygcCnNKwr} {Measuring
  compositional generalization: {A} comprehensive method on realistic data}.
\newblock In \emph{8th International Conference on Learning Representations,
  {ICLR} 2020, Addis Ababa, Ethiopia, April 26-30, 2020}. OpenReview.net.

\bibitem[{Lake(2019)}]{DBLP:conf/nips/Lake19}
Brenden~M. Lake. 2019.
\newblock \href
  {https://proceedings.neurips.cc/paper/2019/hash/f4d0e2e7fc057a58f7ca4a391f01940a-Abstract.html}
  {Compositional generalization through meta sequence-to-sequence learning}.
\newblock In \emph{Advances in Neural Information Processing Systems 32: Annual
  Conference on Neural Information Processing Systems 2019, NeurIPS 2019,
  December 8-14, 2019, Vancouver, BC, Canada}, pages 9788--9798.

\bibitem[{Lake and Baroni(2018)}]{DBLP:conf/icml/LakeB18}
Brenden~M. Lake and Marco Baroni. 2018.
\newblock \href {http://proceedings.mlr.press/v80/lake18a.html} {Generalization
  without systematicity: On the compositional skills of sequence-to-sequence
  recurrent networks}.
\newblock In \emph{Proceedings of the 35th International Conference on Machine
  Learning, {ICML} 2018, Stockholmsm{\"{a}}ssan, Stockholm, Sweden, July 10-15,
  2018}, volume~80 of \emph{Proceedings of Machine Learning Research}, pages
  2879--2888. {PMLR}.

\bibitem[{Lewis et~al.(2020)Lewis, Liu, Goyal, Ghazvininejad, Mohamed, Levy,
  Stoyanov, and Zettlemoyer}]{lewis-etal-2020-bart}
Mike Lewis, Yinhan Liu, Naman Goyal, Marjan Ghazvininejad, Abdelrahman Mohamed,
  Omer Levy, Veselin Stoyanov, and Luke Zettlemoyer. 2020.
\newblock \href {https://doi.org/10.18653/v1/2020.acl-main.703} {{BART}:
  Denoising sequence-to-sequence pre-training for natural language generation,
  translation, and comprehension}.
\newblock In \emph{Proceedings of the 58th Annual Meeting of the Association
  for Computational Linguistics}, pages 7871--7880, Online. Association for
  Computational Linguistics.

\bibitem[{Li et~al.(2019)Li, Zhao, Wang, and
  Hestness}]{DBLP:conf/emnlp/LiZWH19}
Yuanpeng Li, Liang Zhao, Jianyu Wang, and Joel Hestness. 2019.
\newblock \href {https://doi.org/10.18653/v1/D19-1438} {Compositional
  generalization for primitive substitutions}.
\newblock In \emph{Proceedings of the 2019 Conference on Empirical Methods in
  Natural Language Processing and the 9th International Joint Conference on
  Natural Language Processing, {EMNLP-IJCNLP} 2019, Hong Kong, China, November
  3-7, 2019}, pages 4292--4301. Association for Computational Linguistics.

\bibitem[{Liu et~al.(2021)Liu, Yuan, Fu, Jiang, Hayashi, and
  Neubig}]{DBLP:journals/corr/abs-2107-13586}
Pengfei Liu, Weizhe Yuan, Jinlan Fu, Zhengbao Jiang, Hiroaki Hayashi, and
  Graham Neubig. 2021.
\newblock \href {http://arxiv.org/abs/2107.13586} {Pre-train, prompt, and
  predict: {A} systematic survey of prompting methods in natural language
  processing}.
\newblock \emph{CoRR}, abs/2107.13586.

\bibitem[{Nijkamp et~al.(2022)Nijkamp, Pang, Hayashi, Tu, Wang, Zhou, Savarese,
  and Xiong}]{Nijkamp2022ACP}
Erik Nijkamp, Bo~Pang, Hiroaki Hayashi, Lifu Tu, Huan Wang, Yingbo Zhou, Silvio
  Savarese, and Caiming Xiong. 2022.
\newblock A conversational paradigm for program synthesis.
\newblock \emph{arXiv preprint}.

\bibitem[{Qiu et~al.(2022)Qiu, Shaw, Pasupat, Shi, Herzig, Pitler, Sha, and
  Toutanova}]{DBLP:journals/corr/abs-2205-12253}
Linlu Qiu, Peter Shaw, Panupong Pasupat, Tianze Shi, Jonathan Herzig, Emily
  Pitler, Fei Sha, and Kristina Toutanova. 2022.
\newblock \href {https://doi.org/10.48550/arXiv.2205.12253} {Evaluating the
  impact of model scale for compositional generalization in semantic parsing}.
\newblock \emph{CoRR}, abs/2205.12253.

\bibitem[{Raffel et~al.(2020)Raffel, Shazeer, Roberts, Lee, Narang, Matena,
  Zhou, Li, and Liu}]{DBLP:journals/jmlr/RaffelSRLNMZLL20}
Colin Raffel, Noam Shazeer, Adam Roberts, Katherine Lee, Sharan Narang, Michael
  Matena, Yanqi Zhou, Wei Li, and Peter~J. Liu. 2020.
\newblock \href {http://jmlr.org/papers/v21/20-074.html} {Exploring the limits
  of transfer learning with a unified text-to-text transformer}.
\newblock \emph{J. Mach. Learn. Res.}, 21:140:1--140:67.

\bibitem[{Russin et~al.(2019)Russin, Jo, O'Reilly, and
  Bengio}]{DBLP:journals/corr/abs-1904-09708}
Jake Russin, Jason Jo, Randall~C. O'Reilly, and Yoshua Bengio. 2019.
\newblock \href {http://arxiv.org/abs/1904.09708} {Compositional generalization
  in a deep seq2seq model by separating syntax and semantics}.
\newblock \emph{CoRR}, abs/1904.09708.

\bibitem[{Schick and Sch{\"{u}}tze(2021)}]{DBLP:conf/naacl/SchickS21}
Timo Schick and Hinrich Sch{\"{u}}tze. 2021.
\newblock \href {https://doi.org/10.18653/v1/2021.naacl-main.185} {It's not
  just size that matters: Small language models are also few-shot learners}.
\newblock In \emph{Proceedings of the 2021 Conference of the North American
  Chapter of the Association for Computational Linguistics: Human Language
  Technologies, {NAACL-HLT} 2021, Online, June 6-11, 2021}, pages 2339--2352.
  Association for Computational Linguistics.

\bibitem[{Shaw et~al.(2021)Shaw, Chang, Pasupat, and
  Toutanova}]{DBLP:conf/acl/ShawCPT20}
Peter Shaw, Ming{-}Wei Chang, Panupong Pasupat, and Kristina Toutanova. 2021.
\newblock \href {https://doi.org/10.18653/v1/2021.acl-long.75} {Compositional
  generalization and natural language variation: Can a semantic parsing
  approach handle both?}
\newblock In \emph{Proceedings of the 59th Annual Meeting of the Association
  for Computational Linguistics and the 11th International Joint Conference on
  Natural Language Processing, {ACL/IJCNLP} 2021, (Volume 1: Long Papers),
  Virtual Event, August 1-6, 2021}, pages 922--938. Association for
  Computational Linguistics.

\bibitem[{Shin et~al.(2021)Shin, Lin, Thomson, Chen, Roy, Platanios, Pauls,
  Klein, Eisner, and Durme}]{DBLP:conf/emnlp/ShinLTCRPPKED21}
Richard Shin, Christopher~H. Lin, Sam Thomson, Charles Chen, Subhro Roy,
  Emmanouil~Antonios Platanios, Adam Pauls, Dan Klein, Jason Eisner, and
  Benjamin~Van Durme. 2021.
\newblock \href {https://doi.org/10.18653/v1/2021.emnlp-main.608} {Constrained
  language models yield few-shot semantic parsers}.
\newblock In \emph{Proceedings of the 2021 Conference on Empirical Methods in
  Natural Language Processing, {EMNLP} 2021, Virtual Event / Punta Cana,
  Dominican Republic, 7-11 November, 2021}, pages 7699--7715. Association for
  Computational Linguistics.

\bibitem[{Tang and Mooney(2001)}]{DBLP:conf/ecml/TangM01}
Lappoon~R. Tang and Raymond~J. Mooney. 2001.
\newblock \href {https://doi.org/10.1007/3-540-44795-4\_40} {Using multiple
  clause constructors in inductive logic programming for semantic parsing}.
\newblock In \emph{Machine Learning: {EMCL} 2001, 12th European Conference on
  Machine Learning, Freiburg, Germany, September 5-7, 2001, Proceedings},
  volume 2167 of \emph{Lecture Notes in Computer Science}, pages 466--477.
  Springer.

\bibitem[{Tsarkov et~al.(2021)Tsarkov, Tihon, Scales, Momchev, Sinopalnikov,
  and Sch{\"{a}}rli}]{DBLP:conf/aaai/TsarkovTSMSS21}
Dmitry Tsarkov, Tibor Tihon, Nathan Scales, Nikola Momchev, Danila
  Sinopalnikov, and Nathanael Sch{\"{a}}rli. 2021.
\newblock \href {https://ojs.aaai.org/index.php/AAAI/article/view/17195}
  {*-cfq: Analyzing the scalability of machine learning on a compositional
  task}.
\newblock In \emph{Thirty-Fifth {AAAI} Conference on Artificial Intelligence,
  {AAAI} 2021, Thirty-Third Conference on Innovative Applications of Artificial
  Intelligence, {IAAI} 2021, The Eleventh Symposium on Educational Advances in
  Artificial Intelligence, {EAAI} 2021, Virtual Event, February 2-9, 2021},
  pages 9949--9957. {AAAI} Press.

\bibitem[{Wang et~al.(2022{\natexlab{a}})Wang, Roberts, Hesslow, Scao, Chung,
  Beltagy, Launay, and Raffel}]{DBLP:conf/icml/WangRHSCBLR22}
Thomas Wang, Adam Roberts, Daniel Hesslow, Teven~Le Scao, Hyung~Won Chung,
  Iz~Beltagy, Julien Launay, and Colin Raffel. 2022{\natexlab{a}}.
\newblock \href {https://proceedings.mlr.press/v162/wang22u.html} {What
  language model architecture and pretraining objective works best for
  zero-shot generalization?}
\newblock In \emph{International Conference on Machine Learning, {ICML} 2022,
  17-23 July 2022, Baltimore, Maryland, {USA}}, volume 162 of \emph{Proceedings
  of Machine Learning Research}, pages 22964--22984. {PMLR}.

\bibitem[{Wang et~al.(2022{\natexlab{b}})Wang, Li, Xu, Zhou, Lei, Lin, Wang,
  Yang, Zhu, Hoiem, Chang, Bansal, and Ji}]{DBLP:journals/corr/abs-2205-10747}
Zhenhailong Wang, Manling Li, Ruochen Xu, Luowei Zhou, Jie Lei, Xudong Lin,
  Shuohang Wang, Ziyi Yang, Chenguang Zhu, Derek Hoiem, Shih{-}Fu Chang, Mohit
  Bansal, and Heng Ji. 2022{\natexlab{b}}.
\newblock \href {https://doi.org/10.48550/arXiv.2205.10747} {Language models
  with image descriptors are strong few-shot video-language learners}.
\newblock \emph{CoRR}, abs/2205.10747.

\bibitem[{Yin et~al.(2021)Yin, Fang, Neubig, Pauls, Platanios, Su, Thomson, and
  Andreas}]{DBLP:conf/naacl/YinFNPPSTA21}
Pengcheng Yin, Hao Fang, Graham Neubig, Adam Pauls, Emmanouil~Antonios
  Platanios, Yu~Su, Sam Thomson, and Jacob Andreas. 2021.
\newblock \href {https://doi.org/10.18653/v1/2021.naacl-main.225}
  {Compositional generalization for neural semantic parsing via span-level
  supervised attention}.
\newblock In \emph{Proceedings of the 2021 Conference of the North American
  Chapter of the Association for Computational Linguistics: Human Language
  Technologies, {NAACL-HLT} 2021, Online, June 6-11, 2021}, pages 2810--2823.
  Association for Computational Linguistics.

\bibitem[{Zelle and Mooney(1996)}]{DBLP:conf/aaai/ZelleM96}
John~M. Zelle and Raymond~J. Mooney. 1996.
\newblock \href {http://www.aaai.org/Library/AAAI/1996/aaai96-156.php}
  {Learning to parse database queries using inductive logic programming}.
\newblock In \emph{Proceedings of the Thirteenth National Conference on
  Artificial Intelligence and Eighth Innovative Applications of Artificial
  Intelligence Conference, {AAAI} 96, {IAAI} 96, Portland, Oregon, USA, August
  4-8, 1996, Volume 2}, pages 1050--1055. {AAAI} Press / The {MIT} Press.

\bibitem[{Zhang et~al.(2022)Zhang, Roller, Goyal, Artetxe, Chen, Chen, Dewan,
  Diab, Li, Lin, Mihaylov, Ott, Shleifer, Shuster, Simig, Koura, Sridhar, Wang,
  and Zettlemoyer}]{DBLP:journals/corr/abs-2205-01068}
Susan Zhang, Stephen Roller, Naman Goyal, Mikel Artetxe, Moya Chen, Shuohui
  Chen, Christopher Dewan, Mona Diab, Xian Li, Xi~Victoria Lin, Todor Mihaylov,
  Myle Ott, Sam Shleifer, Kurt Shuster, Daniel Simig, Punit~Singh Koura, Anjali
  Sridhar, Tianlu Wang, and Luke Zettlemoyer. 2022.
\newblock \href {https://doi.org/10.48550/arXiv.2205.01068} {{OPT:} open
  pre-trained transformer language models}.
\newblock \emph{CoRR}, abs/2205.01068.

\end{thebibliography}
\newpage
\appendix

\section{Average OOD generalization with respect to average ID generalization performance}
\label{app:ood_v_id}

\begin{figure}[ht]
    \centering
    \includegraphics[width=0.5 \textwidth]{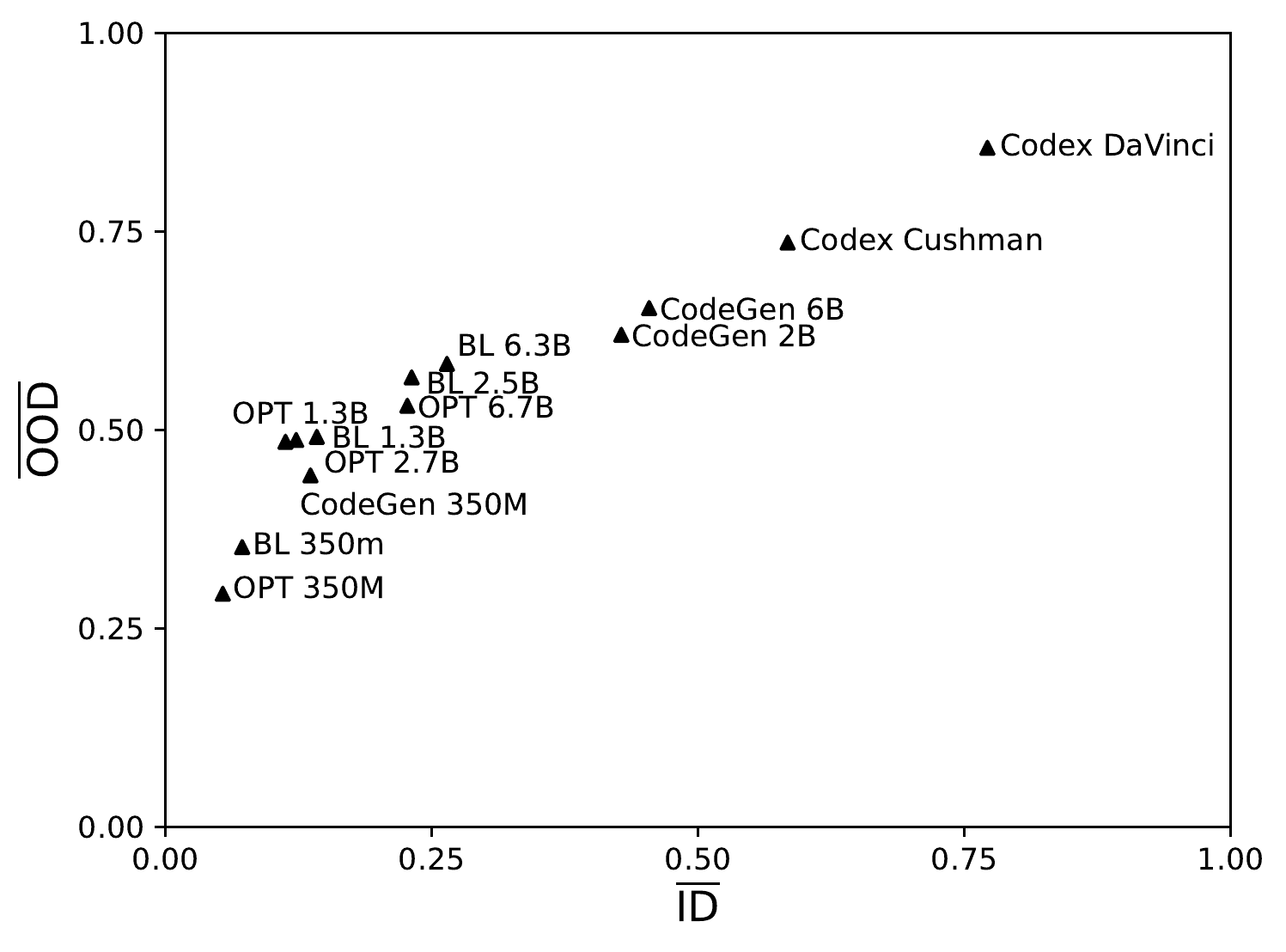}
    \caption{Average OOD generalization vs. average ID generalization performance on GeoQuery-template using 10 exemplars. Results are averaged over five different seeds.}
    \label{fig:geo_ood_v_id}
\end{figure}

\begin{figure}[ht]
    \centering
    \includegraphics[width=0.5 \textwidth]{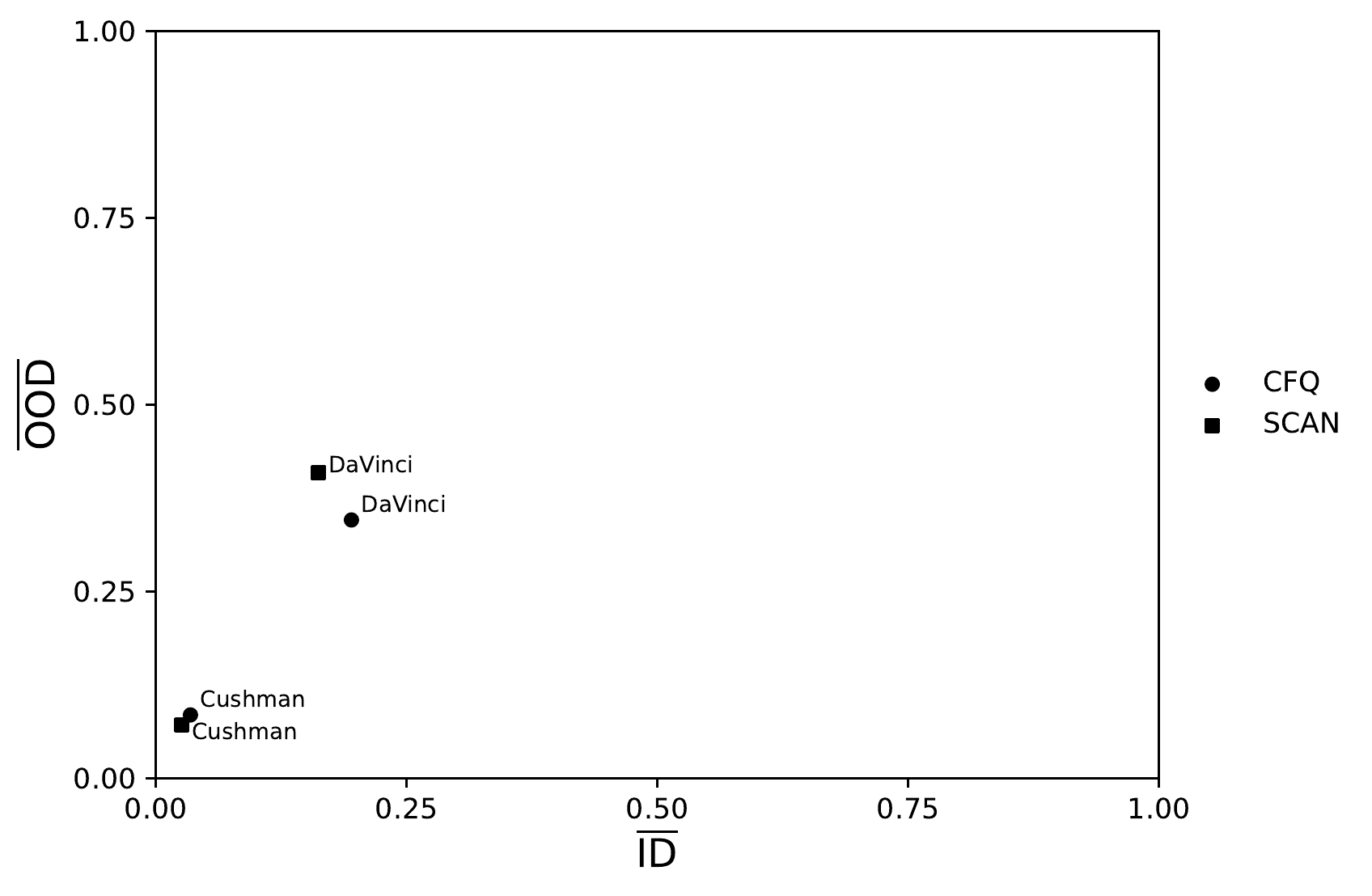}
    \caption{Average OOD generalization vs. average ID generalization performance on CFQ-MCD1 and SCAN-MCD1 using 10 exemplars for Codex DaVinci and Cushman. Results are averaged over five different seeds.}
    \label{fig:cfqscan_ood_v_id}
\end{figure}

\newpage

\pagebreak
\vfill
\section{Prompt design}
\label{app:prompt_design}

Our prompts include a prefix string that introduces the task, followed by a number of input-output examples where inputs and outputs have dataset-specific prefixes. The templates used for producing the prompts are illustrated in Table~\ref{tab:prompt_examples}.

\begin{table*}[ht]
\centering
\small
\begin{tabular}{ll}
\hline & \\[-1.5ex]
\textbf{Dataset} & \textbf{Prompt template}\\
\hline & \\[-1.5ex]
CFQ & \begin{tabular}[c]{@{}l@{}}
\texttt{As a programmer, I can correctly translate any complicated question to a}\\
\texttt{SPARQL query.}\\
\\
\texttt{Question: Was a employer of M1 a film distributor?}\\
\texttt{Query: SELECT count(*) WHERE \{ ?x0 a film.film\_distributor}\\
\texttt{ . ?x0 employment\_tenure.person M1 \}}\\
\\
\texttt{Question: <example 2 input>}\\
\texttt{Query: <example 2 output>}\\
\\
\texttt{...}\\
\\
\texttt{Question: <evaluation input>}\\
\texttt{Query: }
\end{tabular} \\
\hline & \\[-1.5ex]
SCAN & \begin{tabular}[c]{@{}l@{}}
\texttt{Here are some examples of converting complicated commands to correct}\\
\texttt{navigation actions.}\\
\\
\texttt{Command: run opposite right thrice and jump around right thrice.}\\
\texttt{Actions: TURN\_RIGHT TURN\_RIGHT RUN TURN\_RIGHT TURN\_RIGHT RUN}\\
\texttt{ TURN\_RIGHT TURN\_RIGHT RUN TURN\_RIGHT JUMP TURN\_RIGHT JUMP}\\ \texttt{ TURN\_RIGHT JUMP TURN\_RIGHT JUMP TURN\_RIGHT JUMP TURN\_RIGHT JUMP}\\ \texttt{ TURN\_RIGHT JUMP TURN\_RIGHT JUMP TURN\_RIGHT JUMP TURN\_RIGHT}\\
\texttt{ JUMP TURN\_RIGHT JUMP TURN\_RIGHT JUMP.}\\
\\
\texttt{Command: <example 2 input>}\\
\texttt{Actions: <example 2 output>}\\
\\
\texttt{...}\\
\\
\texttt{Command: <evaluation input>}\\
\texttt{Actions: }
\end{tabular} \\
\hline & \\[-1.5ex]
GeoQuery & \begin{tabular}[c]{@{}l@{}}
\texttt{As a programmer, I can correctly translate any complicated question to a}\\
\texttt{meaning representation query.}\\
\\
\texttt{Question: how high is the highest point in m0.}\\
\texttt{Query: answer ( elevation\_1 ( highest ( intersection}\\
\texttt{( place , loc\_2 ( m0 ) ) ) ) ).}\\
\\
\texttt{Question: <example 2 input>}\\
\texttt{Query: <example 2 output>}\\
\\
\texttt{...}\\
\\
\texttt{Question: <evaluation input>}\\
\texttt{Query: }
\end{tabular} \\
\hline
\end{tabular}
\caption{\label{tab:prompt_examples}
Templates used for generating the prompts for CFQ, SCAN, and GeoQuery.
}
\end{table*}

\end{document}